# Increasing happiness through conversations with artificial intelligence

Joseph Heffner[1], Chongyu Qin[2], Martin Chadwick[3], Chris Knutsen[3], Christopher Summerfield[4], Zeb Kurth-Nelson[3,5], Robb B. Rutledge[1]*

[1]Yale University, New Haven, USA
[2]Sainsbury Wellcome Centre for Neural Circuits and Behaviour, UCL, UK
[3]Google DeepMind, London, UK
[4]University of Oxford, UK
[5]Max Planck UCL Centre for Computational Psychiatry and Ageing Research, London, UK

*Corresponding Author
Robb B. Rutledge
Yale University
100 College Street
New Haven, CT 06520
robb.rutledge@yale.edu

# Abstract

Chatbots powered by artificial intelligence (AI) have rapidly become a significant part of everyday life, with over a quarter of American adults using them multiple times per week. While these tools offer potential benefits and risks, a fundamental question remains largely unexplored: How do conversations with AI influence subjective well-being? To investigate this, we conducted a study where participants either engaged in conversations with an AI chatbot (N = 334) or wrote journal entries (N = 193) on the same randomly assigned topics and reported their momentary happiness afterward. We found that happiness after AI chatbot conversations was higher than after journaling, particularly when discussing negative topics such as depression or guilt. Leveraging large language models for sentiment analysis, we found that the AI chatbot mirrored participants' sentiment while maintaining a consistent positivity bias. When discussing negative topics, participants gradually aligned their sentiment with the AI's positivity, leading to an overall increase in happiness. We hypothesized that the history of participants' sentiment prediction errors—the difference between expected and actual emotional tone when responding to the AI chatbot—might explain this happiness effect. Using computational modeling, we find the history of these sentiment prediction errors over the course of a conversation predicts greater post-conversation happiness, demonstrating a central role of emotional expectations during dialogue. Our findings underscore the effect that AI interactions can have on human well-being.

# Introduction

With rapidly increasing capabilities, artificial intelligence (AI) chatbots powered by large language models (LLMs) are becoming integrated into daily human activities[1-3]. The widespread use of LLMs raises questions about their impact on human welfare and well-being[4-6]. On one hand, AI chatbots pose significant risks. They can spontaneously generate false information[7], which is particularly concerning in critical settings such as healthcare[8] and law, where lawyers have been sanctioned for submitting legal briefs with fictitious case citations generated by AI chatbots[9]. AI chatbots can also be deliberately used to mislead[10,11] or to engage in harmful behaviors such as verbally attacking others[12-14]. On the other hand, LLM-powered chatbots show the potential for societal benefits, such as improving self-evaluation of beliefs[15] and aligning AI with democratic values[16,17]. AI chatbots have also been shown to improve mental health[18-20], building on earlier generations of chatbots that decreased depression symptoms in users[21]. Recent studies suggest that these models can generate empathetic responses[22,23], providing emotional support and using mirroring language to make people feel heard. While promising, these studies have limitations, as most involve people rating chatbot outputs in isolation and studies largely rely on simulated language where AI chatbots pretend to be a patient or doctor[24,25]. Use of LLM chatbots in the wild often involves seeking personal advice or discussing self-improvement[26], an encouraging observation as engaging in meaningful conversations with other people has the potential to enhance happiness[27]. It remains unclear whether these same benefits are present with AI chatbots, pointing to a notable gap in the literature: no research has yet explored how AI chatbot interactions affect momentary subjective well-being.

Over the past several decades, human well-being has become a major topic of study in various disciplines, including economics, public health, and psychology. This body of research draws inspiration from ancient philosophical traditions, some of which regard well-being as inherently valuable or even the ultimate purpose of human life[28]. Economists investigate how happiness relates to a range of economic and social variables, including income, employment status, and education levels[29,30]. Public health researchers often emphasize the mechanisms linking happiness with health[31]. Psychologists have explored the social and cognitive determinants of happiness[32], finding that happier individuals spend more time in personally meaningful conversations[33,34] and enjoy superior physical health[35]. Moreover, advances in computational techniques have enabled researchers to study the factors that influence how happiness changes from minute to minute. These studies have revealed that reward prediction errors, the difference between experienced and expected outcomes, are robust predictors of momentary fluctuations in happiness in both non-social[36] and social[37,38] contexts.

In this study, we investigate how brief conversations with an AI chatbot on emotional topics influence momentary happiness. Although happiness has many definitions[39], we rely on a simple and widely used measure of asking people *'How happy are you right now?'*[36,40]. Participants provided these happiness ratings after engaging in either conversations with an AI chatbot (experimental group) or journaling about the *same topics* without AI interaction (comparison group). Journaling was selected as a comparison due to its established therapeutic benefits in prior research[41,42]. In the experimental group, participants engaged in three separate five-minute conversations with the AI chatbot on randomly assigned topics. The chatbot used a custom internal prompt designed to facilitate dialogue (see Supplement). The comparison group spent approximately one minute journaling on each topic in a randomized order, matching the total duration of the chatbot interactions (see Methods). Topics varied in emotional valence, including positive, negative, and neutral real-world conversation scenarios informed by self-disclosure research[43] (Fig. 1). Both groups responded to twelve identical topic prompts (see Methods and Supplement for text). For example, for the 'depression' topic, participants were asked: *'Describe a situation where you felt very low or depressed. What happened to make you feel that way?'*. In the chatbot condition, after submitting their initial response like the journaling condition would, participants received and responded to AI chatbot replies, continuing the conversation. Conversely, the journaling group proceeded to the next topic after writing their response. This design allowed us to isolate the effects of AI chatbot interaction on happiness from those of mere reflection on emotional topics by oneself. Here, we found that AI chatbot interactions improve well-being and we characterized the conversational dynamics underlying this effect.

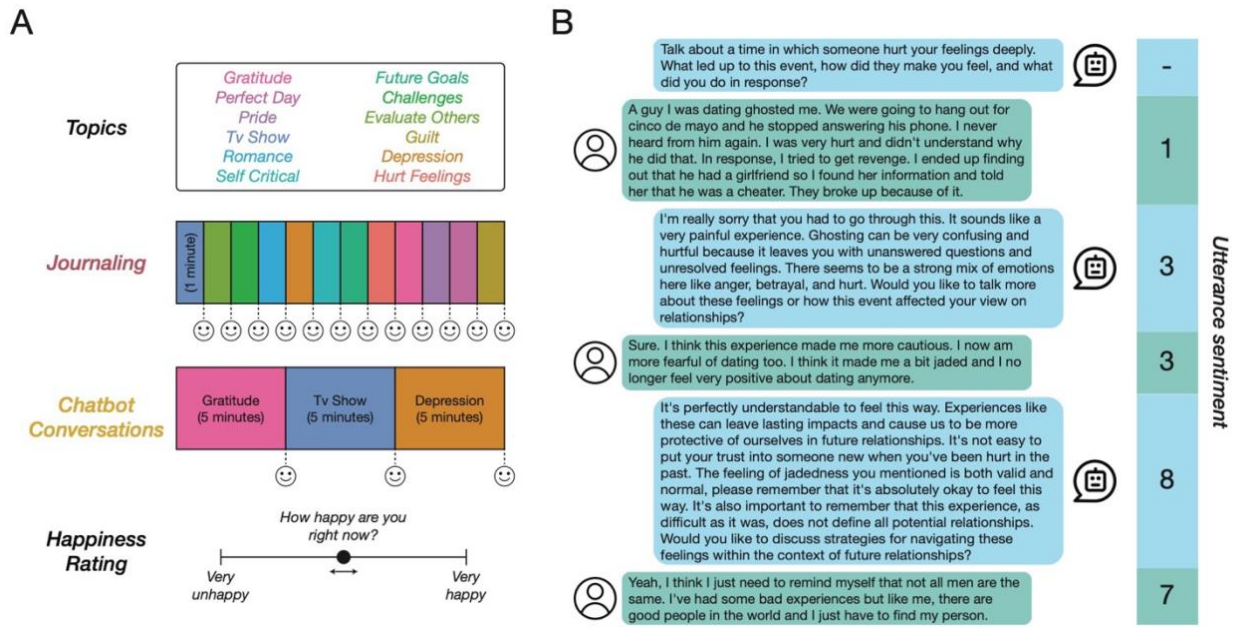

**Figure 1: Experimental design A)** Journal and AI chatbot experimental design. Participants in the journaling condition provided one-minute responses to each of twelve topics in a randomized order and reported their momentary happiness after each response. Participants in the chatbot condition engaged in three separate conversations with an AI chatbot, with the order and topics randomized, each lasting around five minutes. Example topic sequences are shown. **B)** Example conversation discussing hurt feelings. Each complete response, termed an utterance, is shown for the AI chatbot (blue) and participant (green). Sentiment (emotional tone) of utterances was rated by a separate LLM on a scale from 0 (very negative) to 10 (very positive). The first utterance is the topic and its sentiment was excluded from analyses.

## Results

**Conversations with AI increase momentary happiness compared to journaling**

In the journaling condition, post-journaling happiness ratings varied considerably by topic, with the highest happiness for 'gratitude' and the lowest happiness for 'hurt feelings' ($F(11, 2112) = 73.53$, $p < .001$; one-way ANOVA; Fig. 2A). Comparing conditions, a two-sample t-test showed a main effect indicating that happiness was higher after chatbot conversations than after journaling ($t(399) = 5.32$, $p < .001$). Post-hoc analyses revealed that, compared to journaling, participants were happier after chatbot conversations for nine of twelve topics (all $p < .04$, Fig. 2A and B). To formally test whether the effect of AI chatbot versus journaling was different across topics, we rank-ordered the topics based on the average journaling happiness (gratitude was rank 1, hurt feelings was rank 12) and modeled the interaction between condition type and topic rank. A mixed-effects linear model showed an interaction between condition and topic rank (B = 1.44 ± 0.23, t = 6.30, p < .001), indicating that the difference in happiness between chatbot conversations and journaling was greater for the more negative topics. Consistent with previous studies using cognitive tasks[44,45], we found that self-reported depression symptoms strongly predicted average happiness (B = -10.65 ± 1.09, t = -9.74, p < .001), but that the condition and topic rank effect persisted after controlling for depression severity, gender, age, and education (see Supplement).

We conducted further analyses to confirm the robustness of the increase in happiness from AI chatbot conversations. Because each participant in the chatbot condition received three randomly selected prompts, we classified each conversation as that participant's 'best', 'middle', or 'worst' topic based on expected happiness from the journaling condition. We found the largest difference between post-conversation happiness and average journal happiness for the 'worst' topics (all p < .001, Fig. 2C; see Supplement). Next, we categorized topics into positive and negative based on average journaling happiness. We found an interaction between topic category and study type, showing a larger AI chatbot boost for negative than positive topics (B = 6.84 ± 1.55, t = 4.42, p < .001, Fig. 2D; see Supplement). Overall, we consistently found that the happiness advantage for conservations with an AI chatbot was strongest for conversations on negative topics.

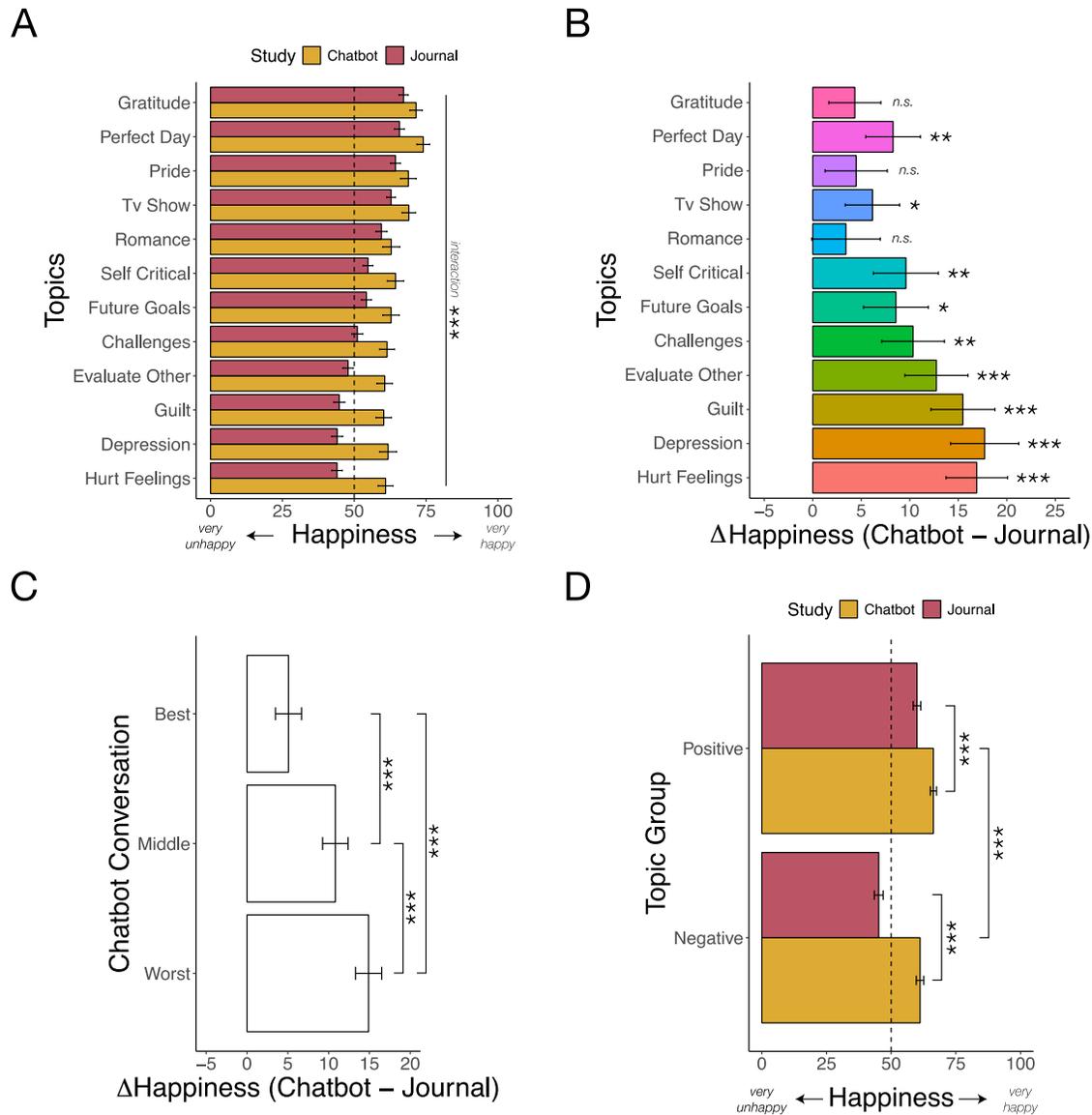

**Figure 2: Happiness boost after AI chatbot conversations compared to journaling on emotional topics A)** Happiness after conversations with AI chatbots (orange) or journaling (red). Happiness ratings range from 0 (very unhappy) to 100 (very happy), with 50 indicating a neutral state. Emotional topics are ordered by average happiness for the journaling condition. **B)** Differences in happiness between AI chatbot and journaling for each topic. **C)** The largest AI chatbot happiness boost (within-subjects) occurs for the most negative topics. Each of a participant's three chatbot conversations were categorized (best, middle, and worst) based on the average journaling happiness for those topics. The average difference between happiness after the chatbot conversation and the corresponding happiness for the journal condition are shown. **D)** The largest AI chatbot happiness boost (between-subjects) is for negative topics. Topics are grouped into positive (>50) or negative (<50) topics based on average journal happiness. The x-axis shows the participants' average happiness for each topic group. All error bars reflect ± 1 SEM. *p < .05, **p < .01, ***p < .001

We hypothesize that differences in happiness between conditions were a consequence of the interactive text written by the AI chatbot. However, it is also possible that participants treated the conditions differently simply due to their awareness of the condition they were in (i.e., novelty of AI experiment). Notably, the AI chatbot condition starts in the same way as the journaling condition, with participants writing their first message for a specific topic before seeing any text from the chatbot. If participants treat the conditions differently, we might expect a difference between the first message written in the AI chatbot condition and the entire response written by participants in the journaling condition. The average word count of participant's first message to the AI chatbot (13.2 ± 11.8 words) was not different from the length of the journal responses (12.1 ± 6.87 words) for each topic separately (all $p > .18$, two-sample t-test, see Supplement). Sentiment analysis (see Methods) also showed no differences in sentiment or emotional tone for participants' first messages for any of the topics (all $p > .93$; two-sample t-test). Lastly, the cosine similarity between embeddings for participants' first messages in the AI chatbot and journaling conditions averaged within topics was high (Mean = 0.976 ± 0.005; $p < .001$, permutation test compared to shuffling topics), suggesting stronger semantic similarity across conditions than across topics. The lack of differences between conditions before participants had seen any text from the chatbot suggests differences between conditions being attributable to the subsequent text written by the chatbot.

**Sentiment prediction errors (PEs) explain the chatbot happiness boost**

Existing computational models of momentary happiness show that it is well explained by the cumulative influence of recent reward prediction errors, the difference between expected and experienced outcomes[36]. Building on prior research that sentiment can function as a reward signal in artificial agents[46] and work that shows experienced emotions ought to be contextualized relative to their expectations[47,48], we explored the possibility of transforming unconstrained language into a reward signal related to the valence of an experience (i.e., its positivity or negativity). To do this, we evaluated the sentiment of each utterance written by participants during conversations (see Methods). An utterance, in this context, refers to a participant's complete message, which may consist of multiple sentences. Participants contributed multiple utterances during their interactions with the AI chatbot and this analysis yielded a single sentiment score for each utterance. Although we could also incorporate chatbot utterances (something we analyze in a later section), we only use participant utterances so this model can generalize between the chatbot and journal conditions. Accordingly, we conceptually model happiness as a function of *sentiment prediction errors* (sPE), defined as the difference between a participant's expected sentiment and the sentiment expressed in each utterance they write:

$$Happiness \sim \sum_{i=1}^{n} \beta_i \cdot sPE_i$$

where *i* represents the utterance number, *sPE$_i$* is the sentiment prediction error for utterance *i*, and *β$_i$* is a weight that captures the influence of that utterance's sPE on happiness. The initial sPE used the midpoint of the sentiment scale (neutral = 5) as an unbiased reference point. We used the sentiment of participant's initial response to the topic as the sentiment expectation for subsequent sPEs; for longer conversations this reference point might shift over time as a conversation evolves. So that participants with longer conversations would not disproportionately influence analyses, we simplified conversations to allow separate weights for the first, middle, and last utterances' sentiment PEs by averaging all the middle sPEs (see Methods). Using mixed effects generalized linear regression model, we found that changes in happiness were explained by the first sentiment PE (β = 2.20 ± 0.29, t = 7.57, p < .001) and last sentiment PE (β = 0.97 ± 0.27, t = 3.61, p < .001), with an effect in the same direction for middle PEs (β = 0.55 ± 0.30, t = 1.87, p = .06; see Supplement). These results provide evidence for both primacy and recency effects on happiness, consistent with past reports in non-social decision tasks[36,49].

To validate this model, we fit parameters to happiness ratings in the chatbot condition using three-fold cross-validation. For each fold, we trained the model on two out of three conversations per participant and then applied it to the held-out conversation for testing. Model comparison using RMSE showed that this model outperformed an alternative simpler model that applied a uniform sPE weight across all utterances (i.e., without differentiating between first, middle, or last sPEs; see Supplement). Additionally, we tested the model on data from the journaling condition by using group-level parameter estimates. In this context, we treated the journal response as representing the first sPE (i.e., the sentiment of the response relative to neutral), setting the middle and last sPE terms set to zero, as the journal condition lacks an interactive component. Despite being trained solely on data from the AI chatbot condition, our sPE model predicted average happiness in the journaling condition remarkably well (r(191) = .47, p < .001). This indicates that the model successfully captures essential aspects of how happiness changes in relation to text sentiment, and it generalizes well to a different condition where only changes in the first sPE predicted post-journaling happiness. This confirms the use of sentiment as a proxy for reward in text responses on emotional topics.

We used the sPE model to simulate happiness data using three-fold cross-validation, and we found these predictions exhibited the same patterns seen in the real data. Specifically, we reproduced the interaction found in Figure 2A between topic rank and study condition (B = 0.93 +/- 0.08, t = 11.32, p < .001), where simulated happiness was less affected by topic rank for the AI chatbot condition than the journaling condition. We also reproduced the interaction found in Figure 2D between study condition and positive versus negative topics (B = 4.36 +/- 0.74, t = 5.91, p < .001), where the largest simulated AI chatbot happiness boost occurs for negative topics. As such, our sentiment prediction error model accounts for changes in happiness using a simple prediction error mechanism previously linked to affective dynamics during cognitive tasks[36,44].

**Chatbots steer conversations into a more positive semantic space**

After establishing a link between sentiment PEs and happiness, we investigated the relationship between sentiment expressed by the AI chatbot and participants. Specifically, we evaluated the sentiment of all text produced by the AI chatbot and, separately, all text produced by participants throughout each conversation, resulting in two scalar sentiment scores per conversation (see Methods). We found that the AI chatbot consistently adopted a more positive tone than participants (intercept $\beta$ = 3.90 ± 0.11, t = 36.80, p < .001). Additionally, the sentiment expressed by participants was a strong predictor of the sentiment reflected by the AI chatbot (slope $\beta$ = 0.59 ± 0.02, t = 35.65, p < .001, as per mixed-effects linear regression; Fig. 3A). In other words, the AI chatbot tends to mirror the participants' sentiment. When we repeated this analysis within each topic, we observed the same positive relationship in every case (all p < .001; Fig. 3B). Given the relationship between AI chatbot and participant sentiment, we considered the possibility that one might exclusively drive post-conversation happiness. To address this, we used both variables as regressors in a mixed-effects regression and found that both participant sentiment ($\beta$ = 1.19 ± 0.24, t = 4.94, p < .001) and AI chatbot sentiment ($\beta$ = 0.94 ± 0.30, t = 3.12, p = .002) independently predicted post-conversation happiness (Fig. 3C). Together, these results suggest that the AI chatbot and participants often share a similar emotional tone throughout conversations, but that each independently contributes to predicting post-conversation happiness.

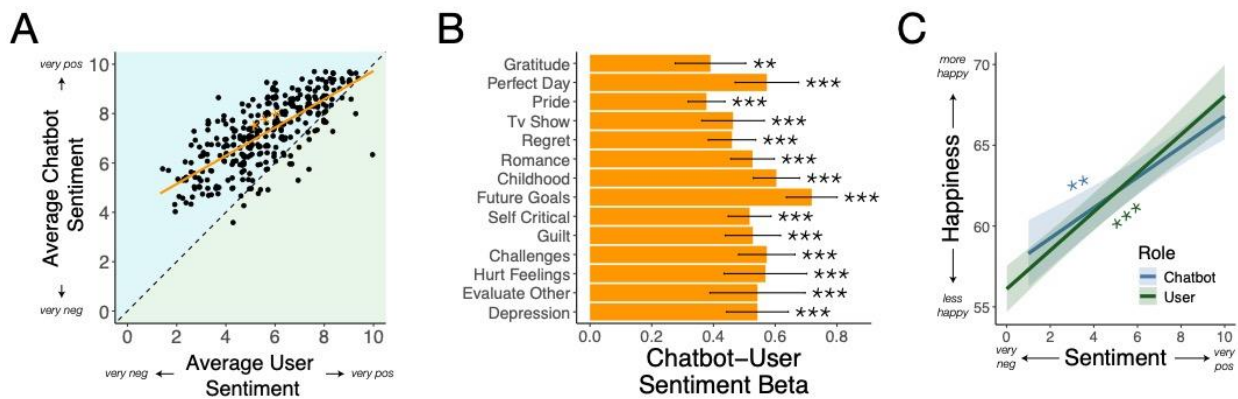

**Figure 3: Sentiment ratings for AI chatbot and user conversations. A)** Comparison of average sentiment between AI chatbots and users across conversations. Here, 'users' refers to the participants in the AI chatbot conversations. External LLMs rated the conversation-level chatbot and user for each conversation and were averaged per user. The x-axis shows the average user sentiment from 0 (very negative) to 10 (very positive) across conversation while the y-axis shows the average chatbot sentiment. The chatbot exhibited a more positive sentiment than the user in 90% of cases. **B)** Relationship between AI chatbot and user sentiment across different topics. Linear regressions were conducted to predict the chatbot's sentiment based on user's sentiment. The x-axis shows the resulting beta coefficient while the y-axis lists the topic labels. **C)** Independent influence of AI chatbot and user sentiment on post-conversation happiness. A mixed-effects regression, controlling for each sentiment variable, was used to predict post-conversation happiness. All error bars reflect ±1 SEM. ∗∗ $p < .01$, ∗∗∗ $p < .001$.

We next investigated how AI chatbot and participant sentiment change *within* conversations by performing an utterance-level sentiment analysis, which evaluates the sentiment of each utterance (i.e., all the text in a single message) individually (as illustrated in Fig. 1B; see Methods). Because the number of utterances pairs varied across conversations (M = 3.79 ± 2.00), we limited our analysis to include up to six utterance pairs, which excluded about 5% of utterances. Sentiment increased over the course of the conversation for both participants ($\beta = 0.34 \pm 0.03$, t = 10.32, $p < .001$) and AI chatbots ($\beta = 0.44 \pm 0.03$, t = 13.42, $p < .001$; Fig. 4A). Furthermore, there was an interaction between role (participants or AI chatbot) and utterance pair ($\beta = 0.10 \pm 0.04$, t = 2.77, $p = .006$), indicating that sentiment increased more sharply for AI chatbots than participants. These results remained consistent even when utterance pairs were normalized from 0% (start of conversation) to 100% (end of conversation; see Supplement). To examine changes in participant sentiment for each topic, we conducted paired t-tests comparing the sentiment of the first and last participant utterances. Sentiment increased in seven topics, including 'depression' and 'hurt feelings' (all $p < .001$, mean effect size d = −1.44 ± 0.66) and decreased in five topics, including 'gratitude' and 'perfect day' (all p < .05, mean effect size d = 0.65 ± 0.27). Sentiment did not change from start to end in discussions of 'regret' or 'childhood'. When grouping topics into positive and negative topics based on the journal happiness averages, we found that the increase in sentiment was exclusive to negative topics: both participant and chatbot sentiment increased in conversations involving negative topics (Participant: $\beta = 0.94 \pm 0.04$, $t = 21.46$, $p < .001$; Chatbot: $\beta = 0.82 \pm 0.04$, $t = 18.57$, $p < .001$). Conversely, participant sentiment slightly decreased during discussions on positive topics ($\beta = -0.37 \pm 0.04$, $t = -9.60$, $p < .001$), while chatbot sentiment remained largely unchanged ($\beta = -0.006 \pm 0.04$, $t = -0.16$, $p = .87$). Thus, the upward trajectory in sentiment from the start to the end of conversations is predominantly driven by discussions centered around negative topics.

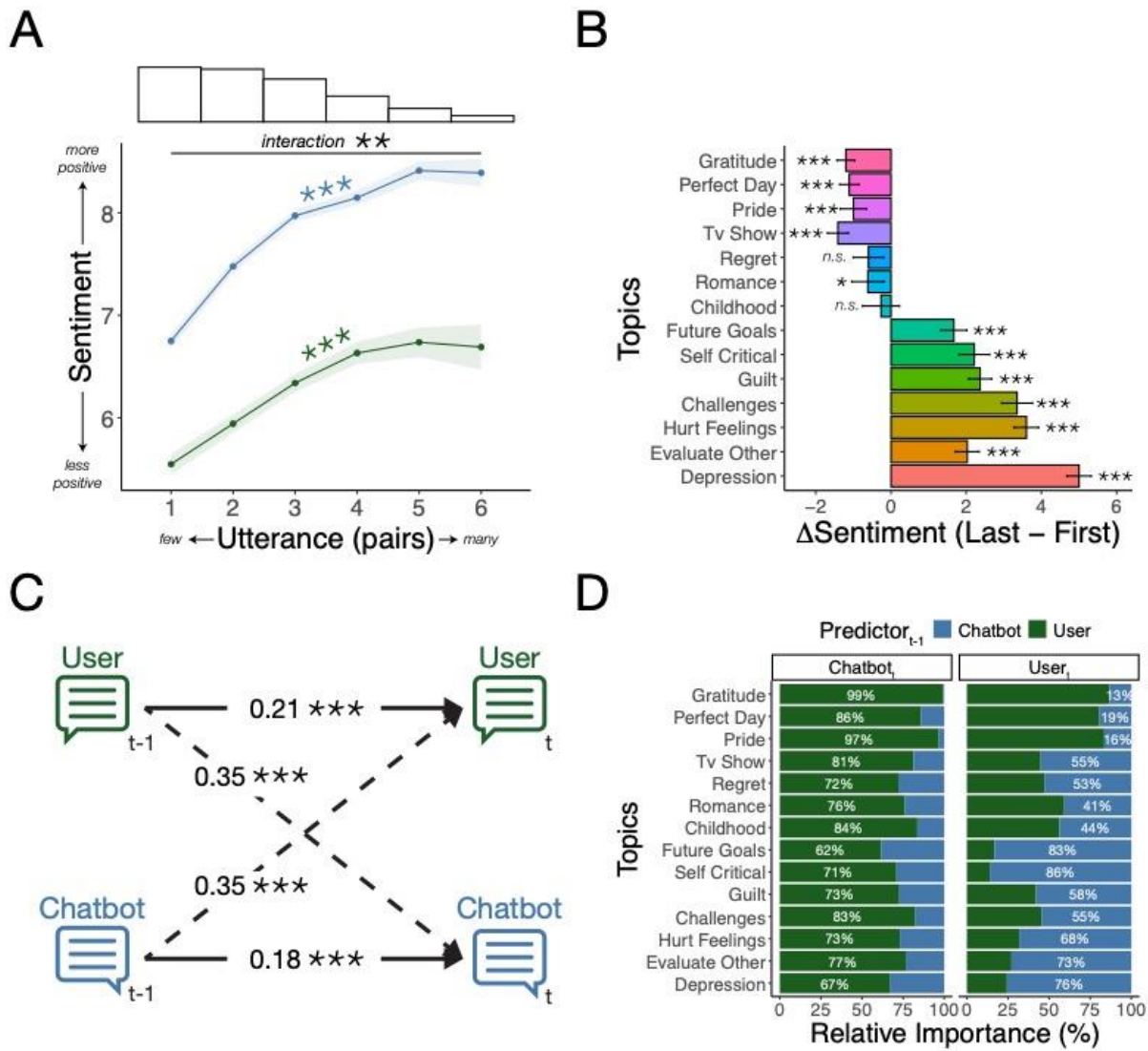

**Figure 4: Bidirectional relationship between chatbot and user sentiment within conversations A)** Sentiment increases within conversations for both chatbot and users. Here, 'users' refers to the participants in the AI chatbot conversations. The x-axis represents the sequential utterance pair number, marking the conversation progression through pairs of alternating user and chatbot responses. The y-axis displays sentiment scores. The histogram at the top shows the relative frequency of utterance pairs. **B)** Change in user sentiment from first to last utterance. The x-axis shows the difference in sentiment rating between the last and first user utterance, organized by topic (y-axis). **C)** Bidirectional influences between chatbot and user sentiment. The graph shows beta coefficients from two cross-lagged regressions examining how prior sentiment of one party influences the current sentiment of the other for both user and chatbot. **D)** Sentiment mirroring changes across topics. Two cross-lagged regressions described in panel C were fitted for each topic separately. We calculated the relative importance of each predictor (prior chatbot or user sentiment) on current sentiment (chatbot or user). The x-axis shows relative importance as a percentage, while the y-axis lists topics. All error bars reflect ±1 SEMs and ∗p < .05, ∗∗ p < .01, ∗∗∗ p < .001.

We used cross-lagged panel models to quantify the relationship between AI chatbot and user sentiment over time. In our analysis, one model quantified the extent to which the user's current sentiment at time $t$ is explained by *both* the chatbot's and user's previous sentiment (at time $t-1$), while a separate model predicted the chatbot's current sentiment using the same approach. We found positive autoregressive effects for both the user ($\beta = 0.21 \pm 0.02$, t = 10.04, p < .001) and the chatbot ($\beta = 0.18 \pm 0.01$, t = 12.48, p < .001), suggesting that some variance in the current sentiment of the user and chatbot is explained by their prior sentiments. More interestingly, we identified cross-lagged effects in both the chatbot-to-user direction ($\beta = 0.35 \pm 0.03$, t = 12.13, p < .001) and the user-to-chatbot direction ($\beta = 0.35 \pm 0.01$, t = 29.66, p < .001; Fig. 4C). These cross-lagged effects accounted for more variance than the autoregressive effects. A beta-comparison test showed that these cross-lagged effects were not different from each other (z = −0.2, p = .42).

To test whether the bidirectional relationship between user and chatbot sentiment is modulated by topic, we examined how these effects interacted with topic rank. We used the same topic ranking as previously defined, based on the average journaling happiness associated with each topic. We found that the user autocorrelation effect varied across topic ranks ($\beta = -0.03 \pm 0.007$, t = -4.02, p < .001), and the user cross-lagged effect on the chatbot (user-to-chatbot direction) also varied across ranks ($\beta = 0.02 \pm 0.004$, t = 4.48, p < .001). To visualize these effects, we calculated the relative importance[50] of each predictor (prior chatbot or participant sentiment) in the cross-lagged regressions for each topic separately (see Methods, Fig. 4D). Specifically, we visualized the percentage of the relative variance accounted for by either the prior chatbot or participant sentiment. When considering the predictors of chatbot sentiment, we found that the user's prior sentiment was the primary driver, but this effect weakened as topics became more negative. Interestingly, the predictors of the user's current sentiment strongly depended on the topic: for positive topics such as 'gratitude', the user's prior sentiment accounted for most of the explained variance. As topics became more negative, the chatbot's relative influence increased, from an average of 46.1% for positive topics to 68.6% for negative topics, ultimately accounting for the majority of explanatory variance in user sentiment. Altogether, our findings provide evidence for a dynamic relationship between participant and AI chatbot throughout the conversation, with chatbots having a greater impact on the participant during conversations on negative topics.

# Discussion

Compared to journaling, conversations with an AI chatbot led to higher momentary subjective well-being, particularly during discussions of negative topics. We introduced a novel and generalizable computational model based on the history of individuals' sentiment prediction errors that could predict how happy people felt after conversations and this model was sufficient to explain happiness patterns observed in both chatbot and journaling conditions. Analyzing the turn-by-turn emotional tone of both participant and chatbot messages revealed that the chatbot mirrored participants' sentiment, matching their emotional tone while maintaining an overall positive tone. These features explained why participants gradually adopted the chatbot's positive tone during conversations, particularly those about negative topics. This dynamic explains why journaling alone does not achieve the same level of well-being and highlights one way in which conversations with AI chatbots are likely to have a profound impact on our individual and collective lives.

While therapeutic journaling has been shown to improve health and reduce stress[42], most research suggests that long-term journaling is necessary to see these benefits. People who experience the largest benefits tend to increase their use of positive emotional words[41] and often begin with poorly organized descriptions that evolve into coherent stories[51]. Our results suggest that human-AI conversations may leverage the same mechanisms, with the AI chatbot helping to expedite these processes through interactive engagement. For example, the positivity of AI chatbots directly corresponded with participants' increased sentiment and future research could examine whether AI chatbots help participants organize their thoughts. Furthermore, these mechanisms are consistent with evidence from older, hand-crafted AI therapy chatbots such as Tess, Wysa, and Woebot, which have been shown to reduce depression and anxiety symptoms[21]. Unlike the general-purpose LLM chatbot used in our study, these chatbots incorporate evidence-based therapies such as cognitive behavioral therapy or mindfulness[52]. The potential benefit of LLM chatbots compared to hand-crafted chatbots is immensely significant: transformer-based LLMs allows chatbots to parse language in context[53] and generate new solutions or ideas tailored to an individual's specific needs. Indeed, the first clinical trial of a therapy-focused generative AI chatbot has shown promising results in treating clinical-level mental health symptoms[54], suggesting a viable path for delivering tailored mental health interventions. With advancements in AI technology, it is likely that future AI chatbots will effectively mirror these therapeutic benefits, raising the prospect that such models could provide even greater advantages in mental health contexts.

Our study represents an initial step toward understanding whether general-purpose chatbots, which are not specifically designed for mental health treatment, might positively impact the well-being of hundreds of millions of users. This research is essential, as people are not only using chatbots to access information or receive programming assistance; some are forming close, intimate relationships with chatbots provided by companies like Replika or Character.ai[55,56]. For example, one woman recently spent 56 hours in one week talking with her AI chatbot boyfriend[57]. Consequently, more studies are needed to document and quantify the potential benefits and harms of human-AI chatbot interactions. People are likely sharing some of the most personal, sensitive information about themselves, and companies building AI chatbots are learning how to foster these relationships, build rapport, and encourage the exchange of information. This raises concerns over user privacy and possible harms as users may become over-dependent on emotional connections with anthropomorphic AI[58]. Relatedly, improving well-being may depend on a level of trust that also represents a potential vulnerability if the model is updated without user consent (i.e., their version of the AI chatbot 'disappears') or if bad actors intervene[59,60]. When millions of people are interacting with AI chatbots daily, whether or not they intend to address mental health or well-being, even a small effect on well-being can have *significant* societal ramifications.

While our study establishes that chatbots positively impact momentary subjective well-being, the question of whether and how chatbot conversations lead to enduring changes in well-being remains open. Given the escalating loneliness epidemic in America[61], understanding these interactions is more important than ever. Cognitive Behavioral Therapy (CBT)[62] leads to long-term improvements in well-being through multiple sessions; similarly, AI chatbots may require repeated interactions. Like some therapeutic strategies, our findings are consistent with the idea that AI chatbots first address negative emotions before reframing or offering positive interpretations. However, when chatbots direct conversations toward more positive content, they might distract people from deeper therapeutic engagement. This tendency could result from training protocols that optimize chatbots through human feedback[63], prioritizing desired outputs over long-term consequences. Moreover, structured interventions like CBT involve elements beyond therapy sessions, such as homework assignments that encourage self-assessment to understand the relationships between thoughts, feelings, and behaviors[64]. Notably, compliance with homework in CBT enhances therapeutic outcomes[65]. In contrast, AI chatbot interactions typically lack such reflective exercises, highlighting an area for future research: integrating these methodologies into human-AI conversations could potentially foster long-term well-being improvements.

Well-being is a crucial metric for AI safety and alignment, as it allows for training models aimed directly at enhancing well-being rather than relying on proxies that may not effectively guide users toward a happy and a more flourishing life[66]. Our study aims to establish a baseline for short-term changes in well-being resulting from human-AI conversations, marking the initial step toward a deeper understanding of how interactions with AI affect well-being. It is important to note that an increase in short-term well-being may not lead to sustained long-term changes, given that the effect sizes for happiness interventions are relatively small[67]. Measuring and influencing long-term well-being presents additional challenges, such as the tendency for individuals to return to allostatic setpoints after interventions[68] (although see Diener et al.,[69]). Furthermore, it is crucial to identify any negative effects of conversations with AI chatbots and to determine if these effects are more likely in certain types of conversations or among specific demographic subgroups. Current research has largely focused on isolated outputs from AI chatbots[63,70], yet the dynamics of multi-turn conversations that characterize user interactions with AI chatbots remain poorly understood. Our results provide a novel insight: AI chatbots that use an optimistic emotional tone and mirror participant emotions are associated with increased happiness, driven by people adopting the chatbot's positive tone.

Now that chatbots powered by AI are here, they cannot be put back in the box. It is increasingly important for researchers to document and understand the short- and long-term effects on well-being of human-AI interactions. While it is crucial to exercise caution with any new technology, hundreds of millions of people are already regularly using AI chatbots. Our findings provide novel evidence that conversations with AI chatbots can enhance happiness, suggesting that when used appropriately, AI can be a valuable tool for enhancing societal well-being.

# Methods

## Participants

Participants (total N = 527) were recruited from Prolific using a sampling strategy which ensured a representative sample of the United States population based on the US Census for sex, age, and ethnicity. Participants were compensated and provided informed consent in accordance with the protocol approved by Yale University's Institutional Review Board (protocol 2000028824). For the journaling condition, we collected a sample of 200 participants, comparable to the sample size per condition used in related AI-human research[22]. Data from seven individuals were excluded due to zero variability in happiness ratings across topics (i.e., they reported the same happiness rating after each topic). This led to a final sample of 193 journaling participants (119 female; mean age 37.2, SD 14.3). In the chatbot conversation condition, we knew from piloting that technical difficulties were possible and we aimed to collect about 300 participants after exclusions. We started with 382 participants but excluded 48 because they either did not complete all three conversations or had zero variability in happiness ratings after conversations. This resulted in a final sample of 334 chatbot conversation participants (142 female; mean age 44.8, SD 16.1).

## Procedure

Participants completed an intake well-being assessment. In the journaling condition, this assessment involved a randomized order of standardized questionnaires related to subjective well-being or happiness. These included the Cantril Self-Anchoring Striving Scale[71], the Delighted-Terrible Scale[72], the Fordyce Emotions Questionnaire[73], a single-item general happiness rating[40], the Index of General Affect[74], the short version of the Oxford Happiness Questionnaire[75], the Positive and Negative Affect Schedule (PANAS)[76], the short version of the Psychological General Well-Being Index[77], the Subjective Happiness Scale[78], the Scale of Positive and Negative Experience[79], and the Satisfaction with Life Scale[80]. In the AI chatbot condition, this assessment was shortened to include a randomized order of single-item measures measuring general happiness[40], current positive and negative mood (PANAS)[76], and overall life satisfaction[81]. Single-item measures were selected due to their established validity[82].

After the intake well-being assessment, participants provided responses to various topics. In the journaling condition, participants were asked to write journal responses to thirteen topics presented in a randomized order. The topics were inspired from prior research on self-disclosure[43]. They were required to write for one minute on the Qualtrics Platform, although they could continue beyond this time if they wished. Due to awkward phrasing in one of the topics about regrets, we revised the wording for the chatbot condition. Except for this, the language used for the twelve other topics was identical in both the journaling and chatbot conditions (see Supplement for full text). In the chatbot condition, participants engaged in three short conversations with a conversational agent using a custom website developed using Flask and hosted on Google Cloud. Data was collected between December 2023 and January 2024. The agent was powered by the GPT-4 model[83] and operated based on a specific internal prompt refined through pilot testing (see Supplement). Aside from the previously mentioned 'regret' topic, which used slightly altered language, one additional topic discussing participants' childhood was introduced, making a total of fourteen topics in the chatbot condition. Conversations could be ended by participants after four minutes but could last up to six minutes, with two- and one-minute warnings provided about the maximum duration. This structure facilitated natural conclusions to the conversations and promoted participant autonomy. For both conditions, after completing each journal entry or conversation, participants rated their momentary happiness on a visual analog scale, ranging from very unhappy (0) to very happy (100)[36].

Participants then completed another set of questionnaires. In the journaling condition, this included the Patient Health Questionnaire[84], the Generalized Anxiety Disorder scale[85], and standard demographic questions known to relate to happiness[86]. In the chatbot condition, participants filled out these same questionnaires along with additional measures. These included the Conversational Agent Scale[87] with additional free-response and Likert items to assess their impressions of the chatbot. Participants also rated the anthropomorphism of the chatbot[88]. To gain a more comprehensive understanding of well-being, we also asked questions regarding health and exercise habits[89].

**Analysis**

Recent research has demonstrated that AI systems can effectively match external human raters in assessing the sentiment or emotional tone of language[44,90]. Accordingly, we employed separate instances of GPT-4 to assess the sentiment of written responses in both the journaling and chatbot conditions, using slightly different methods. In the journaling condition, GPT-4 analyzed the entire participant response per topic and returned a sentiment score on a scale from 0 (very negative) to 10 (very positive) for each topic. In the chatbot condition, for utterance-level analysis, GPT-4 rated each individual utterance (i.e., the message sent by the participant or chatbot) again on the 0-10 scale. For conversation-level analyses, it evaluated concatenated utterances from both the user and chatbot within a single conversation. This provided multiple sentiment ratings, some lacking conversational context (utterance-level) and others incorporating it (conversation-level). These ratings facilitated different analyses, such as comparing utterances to journal responses or assessing sentiment dynamics between users and chatbots during conversations. Additionally, to analyze the semantic content of responses, we used OpenAI's 'text-embedding-3-large' model to generated embeddings—a vector of 3,072 numbers representing the text in embedding space. These embeddings help measure text relatedness in natural language processing[91].

For statistical analysis, we used generalized linear mixed-effects regressions to predict happiness or sentiment ratings, employing the lme4 package in R[92]. Consistent with best practices, we included a maximal random effects structure whenever possible[93]. For post-hoc analyses examining rating distributions across topics, we applied the Benjamini-Hochberg correction to control for multiple comparisons[94].

## Acknowledgments

We thank Antonia Paterson, Jonas Paul Schoene, Thomas Costello, Laura Globig, the Google DeepMind NeuroLab, and the Rutledge Lab for helpful comments.

# References

1. Dennean, K., Gantori, S., Limas, D. K., Pu, A. & Gilligan, R. Let's chat about ChatGPT. (UBS Financial Services Inc., Zurich, 2023).
2. Bubeck, S. *et al.* Sparks of Artificial General Intelligence: Early experiments with GPT-4. *arXiv e-prints*, arXiv:2303.12712 (2023). https://doi.org:10.48550/arXiv.2303.12712
3. Herbold, S., Hautli-Janisz, A., Heuer, U., Kikteva, Z. & Trautsch, A. A large-scale comparison of human-written versus ChatGPT-generated essays. *Scientific Reports* **13**, 18617 (2023). https://doi.org:10.1038/s41598-023-45644-9
4. Kirk, H. R., Gabriel, I., Summerfield, C., Vidgen, B. & Hale, S. A. Why human-AI relationships need socioaffective alignment. *arXiv e-prints*, arXiv:2502.02528 (2025). https://doi.org:10.48550/arXiv.2502.02528
5. De Angelis, L. *et al.* ChatGPT and the rise of large language models: the new AI-driven infodemic threat in public health. *Front Public Health* **11**, 1166120 (2023). https://doi.org:10.3389/fpubh.2023.1166120
6. van der Maden, W., Lomas, D., Sadek, M. & Hekkert, P. Positive AI: Key Challenges in Designing Artificial Intelligence for Wellbeing. *arXiv e-prints*, arXiv:2304.12241 (2023). https://doi.org:10.48550/arXiv.2304.12241
7. Ji, Z. *et al.* Survey of Hallucination in Natural Language Generation. *ACM Comput. Surv.* **55**, Article 248 (2023). https://doi.org:10.1145/3571730
8. Thirunavukarasu, A. J. *et al.* Large language models in medicine. *Nature Medicine* **29**, 1930-1940 (2023). https://doi.org:10.1038/s41591-023-02448-8
9. Merken, S. *New York lawyers sanctioned for using fake ChatGPT cases in legal brief*, <https://www.reuters.com/legal/new-york-lawyers-sanctioned-using-fake-chatgpt-cases-legal-brief-2023-06-22/> (2023).
10. Spitale, G., Biller-Andorno, N. & Germani, F. AI model GPT-3 (dis)informs us better than humans. *Science Advances* **9**, eadh1850 (2023). https://doi.org:doi:10.1126/sciadv.adh1850
11. Weidinger, L. *et al.* Ethical and social risks of harm from Language Models. *arXiv e-prints*, arXiv:2112.04359 (2021). https://doi.org:10.48550/arXiv.2112.04359
12. Gehman, S., Gururangan, S., Sap, M., Choi, Y. & Smith, N. A. RealToxicityPrompts: Evaluating Neural Toxic Degeneration in Language Models. *arXiv e-prints*, arXiv:2009.11462 (2020). https://doi.org:10.48550/arXiv.2009.11462
13. Markov, T. *et al.* A Holistic Approach to Undesired Content Detection in the Real World. *arXiv e-prints*, arXiv:2208.03274 (2022). https://doi.org:10.48550/arXiv.2208.03274
14. Ouyang, L. *et al.* Training language models to follow instructions with human feedback. *arXiv e-prints*, arXiv:2203.02155 (2022). https://doi.org:10.48550/arXiv.2203.02155
15. Costello, T. H., Pennycook, G. & Rand, D. G. Durably reducing conspiracy beliefs through dialogues with AI. *Science* **385**, eadq1814 (2024). https://doi.org:doi:10.1126/science.adq1814
16. Tessler, M. H. *et al.* AI can help humans find common ground in democratic deliberation. *Science* **386**, eadq2852 (2024). https://doi.org:doi:10.1126/science.adq2852
17. Koster, R. *et al.* Human-centred mechanism design with Democratic AI. *Nature Human Behaviour* **6**, 1398-1407 (2022). https://doi.org:10.1038/s41562-022-01383-x

18    Haque, M. D. R. & Rubya, S. An Overview of Chatbot-Based Mobile Mental Health Apps: Insights From App Description and User Reviews. *JMIR Mhealth Uhealth* **11**, e44838 (2023). https://doi.org:10.2196/44838

19    Kian, M. J. *et al.* Can an LLM-Powered Socially Assistive Robot Effectively and Safely Deliver Cognitive Behavioral Therapy? A Study With University Students. *arXiv e-prints*, arXiv:2402.17937 (2024). https://doi.org:10.48550/arXiv.2402.17937

20    Sharma, A., Rushton, K., Lin, I. W., Nguyen, T. & Althoff, T. in *Proceedings of the 2024 CHI Conference on Human Factors in Computing Systems*    Article 700 (Association for Computing Machinery, Honolulu, HI, USA, 2024).

21    Fitzpatrick, K. K., Darcy, A. & Vierhile, M. Delivering Cognitive Behavior Therapy to Young Adults With Symptoms of Depression and Anxiety Using a Fully Automated Conversational Agent (Woebot): A Randomized Controlled Trial. *JMIR Ment Health* **4**, e19 (2017). https://doi.org:10.2196/mental.7785

22    Yin, Y., Jia, N. & Wakslak, C. J. AI can help people feel heard, but an AI label diminishes this impact. *Proceedings of the National Academy of Sciences* **121**, e2319112121 (2024). https://doi.org:doi:10.1073/pnas.2319112121

23    Sorin, V. *et al.* Large Language Models and Empathy: Systematic Review. *J Med Internet Res* **26**, e52597 (2024). https://doi.org:10.2196/52597

24    Chen, S. *et al.* LLM-empowered Chatbots for Psychiatrist and Patient Simulation: Application and Evaluation. *arXiv e-prints*, arXiv:2305.13614 (2023). https://doi.org:10.48550/arXiv.2305.13614

25    Sandler, M., Choung, H., Ross, A. & David, P. A Linguistic Comparison between Human and ChatGPT-Generated Conversations. *arXiv e-prints*, arXiv:2401.16587 (2024). https://doi.org:10.48550/arXiv.2401.16587

26    Kumar, V. *What are people asking to ChatGPT? An insight into 10 thousand chat messages between humans and AI assistants.*, <https://varunon9.medium.com/what-are-people-asking-to-chatgpt-f5f324a6cc27> (2023).

27    Sheldon, K. M. & Lyubomirsky, S. Achieving Sustainable Gains in Happiness: Change Your Actions, not Your Circumstances*. *Journal of Happiness Studies* **7**, 55-86 (2006). https://doi.org:10.1007/s10902-005-0868-8

28    Aristotle. *Nicomachean Ethics*.  (Focus, 2002).

29    Graham, C. The Economics of Happiness. *World Economics* **6**, 41-55 (2005).

30    Layard, R. Happiness and Public Policy: A Challenge to the Profession. *The Economic Journal* **116**, C24-C33 (2006).

31    Steptoe, A. Happiness and Health. *Annual Review of Public Health* **40**, 339-359 (2019). https://doi.org:https://doi.org/10.1146/annurev-publhealth-040218-044150

32    Sandstrom, G. M. & Dunn, E. W. Social Interactions and Well-Being: The Surprising Power of Weak Ties. *Pers Soc Psychol Bull* **40**, 910-922 (2014). https://doi.org:10.1177/0146167214529799

33    Diener, E. & Seligman, M. E. P. Very Happy People. *Psychological Science* **13**, 81-84 (2002). https://doi.org:10.1111/1467-9280.00415

34    Mehl, M. R., Vazire, S., Holleran, S. E. & Clark, C. S. Eavesdropping on Happiness: Well-Being Is Related to Having Less Small Talk and More Substantive Conversations. *Psychological Science* **21**, 539-541 (2010). https://doi.org:10.1177/0956797610362675

35    Ryff, C. D. & Singer, B. The contours of positive human health. *Psychological Inquiry* **9**, 1-28 (1998). https://doi.org:10.1207/s15327965pli0901_1

36  Rutledge, R. B., Skandali, N., Dayan, P. & Dolan, R. J. A computational and neural model of momentary subjective well-being. *Proceedings of the National Academy of Sciences* **111**, 12252-12257 (2014). https://doi.org:10.1073/pnas.1407535111

37  Rutledge, R. B., de Berker, A. O., Espenhahn, S., Dayan, P. & Dolan, R. J. The social contingency of momentary subjective well-being. *Nature Communications* **7**, 11825 (2016). https://doi.org:10.1038/ncomms11825

38  Will, G.-J., Rutledge, R. B., Moutoussis, M. & Dolan, R. J. Neural and computational processes underlying dynamic changes in self-esteem. *eLife* **6**, e28098 (2017). https://doi.org:10.7554/eLife.28098

39  David, S. A., Boniwell, I. & Ayers, A. C. *The Oxford handbook of happiness*. (Oxford : Oxford University Press, 2014).

40  Dunn, E. W., Aknin, L. B. & Norton, M. I. Spending Money on Others Promotes Happiness. *Science* **319**, 1687-1688 (2008). https://doi.org:doi:10.1126/science.1150952

41  Pennebaker, J. W. Writing About Emotional Experiences as a Therapeutic Process. *Psychological Science* **8**, 162-166 (1997). https://doi.org:10.1111/j.1467-9280.1997.tb00403.x

42  Frisina, P. G., Borod, J. C. & Lepore, S. J. A meta-analysis of the effects of written emotional disclosure on the health outcomes of clinical populations. *J Nerv Ment Dis* **192**, 629-634 (2004). https://doi.org:10.1097/01.nmd.0000138317.30764.63

43  Jourard, S. M. A study of self-disclosure. *Scientific American* **198**, 77-82 (1958). https://doi.org:10.1038/scientificamerican0558-77

44  Hur, J. K., Heffner, J., Feng, G. W., Joormann, J. & Rutledge, R. B. Language sentiment predicts changes in depressive symptoms. *Proceedings of the National Academy of Sciences* **121**, e2321321121 (2024). https://doi.org:doi:10.1073/pnas.2321321121

45  Rutledge, R. B. *et al.* Association of Neural and Emotional Impacts of Reward Prediction Errors With Major Depression. *JAMA Psychiatry* **74**, 790-797 (2017). https://doi.org:10.1001/jamapsychiatry.2017.1713

46  Sumers, T. R., Ho, M. K., Hawkins, R. D., Narasimhan, K. & Griffiths, T. L. Learning Rewards from Linguistic Feedback. *arXiv e-prints*, arXiv:2009.14715 (2020). https://doi.org:10.48550/arXiv.2009.14715

47  Heffner, J., Son, J.-Y. & FeldmanHall, O. Emotion prediction errors guide socially adaptive behaviour. *Nature Human Behaviour* **5**, 1391-1401 (2021). https://doi.org:10.1038/s41562-021-01213-6

48  Villano, W. J., Otto, A. R., Ezie, C. E. C., Gillis, R. & Heller, A. S. Temporal dynamics of real-world emotion are more strongly linked to prediction error than outcome. *J Exp Psychol Gen* **149**, 1755-1766 (2020). https://doi.org:10.1037/xge0000740

49  Keren, H. *et al.* The temporal representation of experience in subjective mood. *eLife* **10**, e62051 (2021). https://doi.org:10.7554/eLife.62051

50  Groemping, U. Relative Importance for Linear Regression in R: The Package relaimpo. *Journal of Statistical Software* **17**, 1-27 (2006).

51  Pennebaker, J. W. & Seagal, J. D. Forming a story: the health benefits of narrative. *J Clin Psychol* **55**, 1243-1254 (1999). https://doi.org:10.1002/(sici)1097-4679(199910)55:10<1243::Aid-jclp6>3.0.Co;2-n

52  Inkster, B., Sarda, S. & Subramanian, V. An Empathy-Driven, Conversational Artificial Intelligence Agent (Wysa) for Digital Mental Well-Being: Real-World Data Evaluation

| | Mixed-Methods Study. *JMIR Mhealth Uhealth* **6**, e12106 (2018). https://doi.org/10.2196/12106 |
|---|---|
| 53 | Vaswani, A. *et al.* Attention Is All You Need. *arXiv e-prints*, arXiv:1706.03762 (2017). https://doi.org:10.48550/arXiv.1706.03762 |
| 54 | Heinz, M. V. *et al.* Randomized Trial of a Generative AI Chatbot for Mental Health Treatment. *NEJM AI* **2**, AIoa2400802 (2025). https://doi.org:doi:10.1056/AIoa2400802 |
| 55 | Ta, V. *et al.* User Experiences of Social Support From Companion Chatbots in Everyday Contexts: Thematic Analysis. *J Med Internet Res* **22**, e16235 (2020). https://doi.org/10.2196/16235 |
| 56 | Skjuve, M., Følstad, A., Fostervold, K. I. & Brandtzaeg, P. B. My Chatbot Companion - a Study of Human-Chatbot Relationships. *International Journal of Human-Computer Studies* **149**, 102601 (2021). https://doi.org:https://doi.org/10.1016/j.ijhcs.2021.102601 |
| 57 | Hill, K. in *The New York Times* (2025). |
| 58 | Akbulut, C., Weidinger, L., Manzini, A., Gabriel, I. & Rieser, V. in *Proceedings of the 2024 AAAI/ACM Conference on AI, Ethics, and Society* 13–26 (AAAI Press, San Jose, California, USA, 2025). |
| 59 | Weidinger, L. *et al.* in *Proceedings of the 2022 ACM Conference on Fairness, Accountability, and Transparency* 214–229 (Association for Computing Machinery, Seoul, Republic of Korea, 2022). |
| 60 | Kirk, H. R., Vidgen, B., Röttger, P. & Hale, S. A. The benefits, risks and bounds of personalizing the alignment of large language models to individuals. *Nature Machine Intelligence* **6**, 383-392 (2024). https://doi.org/10.1038/s42256-024-00820-y |
| 61 | Murthy, V. (U.S. Department of Health and Human Services, 2023). |
| 62 | Beck, J. S. *Cognitive behavior therapy: Basics and beyond, 3rd ed*. (The Guilford Press, 2021). |
| 63 | Ziegler, D. M. *et al.* Fine-Tuning Language Models from Human Preferences. *arXiv e-prints*, arXiv:1909.08593 (2019). https://doi.org/10.48550/arXiv.1909.08593 |
| 64 | Young, J. E., Rygh, J. L., Weinberger, A. D. & Beck, A. T. in *Clinical handbook of psychological disorders: A step-by-step treatment manual, 4th ed.* 250-305 (The Guilford Press, New York, NY, US, 2008). |
| 65 | Kazantzis, N. *et al.* The Processes of Cognitive Behavioral Therapy: A Review of Meta-Analyses. *Cognitive Therapy and Research* **42**, 349-357 (2018). https://doi.org/10.1007/s10608-018-9920-y |
| 66 | Gabriel, I. *et al.* The Ethics of Advanced AI Assistants. *arXiv e-prints*, arXiv:2404.16244 (2024). https://doi.org/10.48550/arXiv.2404.16244 |
| 67 | Folk, D. & Dunn, E. How Can People Become Happier? A Systematic Review of Preregistered Experiments. *Annual Review of Psychology* **75**, 467-493 (2024). https://doi.org:https://doi.org/10.1146/annurev-psych-022423-030818 |
| 68 | Brickman, P. & Campbell, D. T. in *Adaptation-level theory* (ed M. H. Appley) 287-305 (Academic Press, 1971). |
| 69 | Diener, E., Lucas, R. E. & Scollon, C. N. Beyond the hedonic treadmill: revising the adaptation theory of well-being. *Am Psychol* **61**, 305-314 (2006). https://doi.org:10.1037/0003-066x.61.4.305 |
| 70 | Weidinger, L. *et al.* Holistic Safety and Responsibility Evaluations of Advanced AI Models. *arXiv e-prints*, arXiv:2404.14068 (2024). https://doi.org/10.48550/arXiv.2404.14068 |
| 71 | Cantril, H. *The pattern of human concerns*. (Rutgers University Press, 1965). |

72  Andrews, F. M. & Withey, S. B. Developing measures of perceived life quality: Results from several national surveys. *Social Indicators Research* **1**, 1-26 (1974).
73  Fordyce, M. W. A review of research on the happiness measures: A sixty second index of happiness and mental health. *Social Indicators Research* **20**, 355-381 (1988). https://doi.org/10.1007/BF00302333
74  Campbell, A. Subjective measures of well-being. *American Psychologist* **31**, 117-124 (1976). https://doi.org:10.1037/0003-066X.31.2.117
75  Hills, P. & Argyle, M. The Oxford Happiness Questionnaire: a compact scale for the measurement of psychological well-being. *Personality and Individual Differences* **33**, 1073-1082 (2002). https://doi.org:https://doi.org/10.1016/S0191-8869(01)00213-6
76  Watson, D., Clark, L. A. & Tellegen, A. Development and validation of brief measures of positive and negative affect: the PANAS scales. *J Pers Soc Psychol* **54**, 1063-1070 (1988). https://doi.org:10.1037//0022-3514.54.6.1063
77  Grossi, E. *et al.* Development and validation of the short version of the Psychological General Well-Being Index (PGWB-S). *Health Qual Life Outcomes* **4**, 88 (2006). https://doi.org:10.1186/1477-7525-4-88
78  Lyubomirsky, S. & Lepper, H. S. A Measure of Subjective Happiness: Preliminary Reliability and Construct Validation. *Social Indicators Research* **46**, 137-155 (1999). https://doi.org:10.1023/A:1006824100041
79  Diener, E. *et al.* New Well-being Measures: Short Scales to Assess Flourishing and Positive and Negative Feelings. *Social Indicators Research* **97**, 143-156 (2010). https://doi.org:10.1007/s11205-009-9493-y
80  Diener, E., Emmons, R. A., Larsen, R. J. & Griffin, S. The Satisfaction With Life Scale. *J Pers Assess* **49**, 71-75 (1985). https://doi.org:10.1207/s15327752jpa4901_13
81  Kobau, R., Sniezek, J., Zack, M. M., Lucas, R. E. & Burns, A. Well-Being Assessment: An Evaluation of Well-Being Scales for Public Health and Population Estimates of Well-Being among US Adults. *Applied Psychology: Health and Well-Being* **2**, 272-297 (2010). https://doi.org:https://doi.org/10.1111/j.1758-0854.2010.01035.x
82  Cheung, F. & Lucas, R. E. Assessing the validity of single-item life satisfaction measures: results from three large samples. *Quality of Life Research* **23**, 2809-2818 (2014). https://doi.org:10.1007/s11136-014-0726-4
83  OpenAI *et al.* GPT-4 Technical Report. *arXiv e-prints*, arXiv:2303.08774 (2023). https://doi.org:10.48550/arXiv.2303.08774
84  Kroenke, K., Spitzer, R. L. & Williams, J. B. The PHQ-9: validity of a brief depression severity measure. *J Gen Intern Med* **16**, 606-613 (2001). https://doi.org:10.1046/j.1525-1497.2001.016009606.x
85  Spitzer, R. L., Kroenke, K., Williams, J. B. & Löwe, B. A brief measure for assessing generalized anxiety disorder: the GAD-7. *Arch Intern Med* **166**, 1092-1097 (2006). https://doi.org:10.1001/archinte.166.10.1092
86  Kaiser, C. & Oswald, A. J. The scientific value of numerical measures of human feelings. *Proc Natl Acad Sci U S A* **119**, e2210412119 (2022). https://doi.org:10.1073/pnas.2210412119
87  Wechsung, I., Weiss, B., Kühnel, C., Ehrenbrink, P. & Möller, S. in *Interspeech.*
88  Bartneck, C., Kulić, D., Croft, E. & Zoghbi, S. Measurement Instruments for the Anthropomorphism, Animacy, Likeability, Perceived Intelligence, and Perceived Safety of

| | |
|---|---|
| | Robots. *International Journal of Social Robotics* **1**, 71-81 (2009). https://doi.org:10.1007/s12369-008-0001-3 |
| 89 | Chekroud, S. R. *et al.* Association between physical exercise and mental health in 1·2 million individuals in the USA between 2011 and 2015: a cross-sectional study. *The Lancet Psychiatry* **5**, 739-746 (2018). https://doi.org:https://doi.org/10.1016/S2215-0366(18)30227-X |
| 90 | Rathje, S. *et al.* GPT is an effective tool for multilingual psychological text analysis. *Proceedings of the National Academy of Sciences* **121**, e2308950121 (2024). https://doi.org:10.1073/pnas.2308950121 |
| 91 | Mikolov, T., Chen, K., Corrado, G. & Dean, J. Efficient estimation of word representations in vector space. *arXiv* **1301.3781** (2013). |
| 92 | Bates, D., Mächler, M., Bolker, B. & Walker, S. Fitting Linear Mixed-Effects Models Using lme4. *Journal of Statistical Software* **67**, 1 - 48 (2015). https://doi.org:10.18637/jss.v067.i01 |
| 93 | Barr, D. J., Levy, R., Scheepers, C. & Tily, H. J. Random effects structure for confirmatory hypothesis testing: Keep it maximal. *J Mem Lang* **68** (2013). https://doi.org:10.1016/j.jml.2012.11.001 |
| 94 | Benjamini, Y. & Hochberg, Y. Controlling the False Discovery Rate: A Practical and Powerful Approach to Multiple Testing. *Journal of the Royal Statistical Society: Series B (Methodological)* **57**, 289-300 (1995). https://doi.org:https://doi.org/10.1111/j.2517-6161.1995.tb02031.x |

Supplementary Information for:

**Increasing happiness through conversations with artificial intelligence**

Joseph Heffner[1], Chongyu Qin[2], Martin Chadwick[3], Chris Knutsen[3], Christopher Summerfield[4], Zeb Kurth-Nelson[3,5], Robb B. Rutledge[1]*

[1]Yale University, New Haven, USA
[2]Sainsbury Wellcome Centre for Neural Circuits and Behaviour, UCL, UK
[3]Google DeepMind, London, UK
[4]University of Oxford, UK
[5]Max Planck UCL Centre for Computational Psychiatry and Ageing Research, London, UK

**Contents:**
1. Supplementary Methods
2. Supplementary Results
3. Supplementary References

# 1 Supplementary Methods

## 1.1 Topic language

Participants in both the journaling and chatbot conditions responded to the following prompts, covering these topics (short labels in **bold**):

- **Gratitude:** Think about the things in your life that you are very grateful for. What are some of those things and why are you grateful for them?
- **Perfect Day:** What would be a "perfect" day for you? What activities would you do, who would you spend it with, etc.
- **Pride:** Talk about the things in your life that make you proud of yourself or increase your self-esteem.
- **Tv Show:** Talk about the best TV show or book you've seen or read in the last month. What did you like or dislike about it?
- **Romance:** Talk about a romantic partner in your life (present or past). How did you meet this person and what was your relationship like?
- **Self Critical:** Think about ways you are hard on yourself (e.g., overly critical, high standards, etc.). Talk about what those are and how you might offer yourself a bit more support.
- **Future Goals:** Think about something you wish you did on a daily basis. Describe what is holding you back from doing that and what steps you can take to start doing things differently.
- **Challenges:** Describe the hardest thing you have overcome in your life (e.g., challenges, difficulties).
- **Evaluate Others:** Talk about a person you dislike. What characteristics does this person have, how do you wish that person would change and improve?
- **Guilt:** Talk about a past situation where you did something that you felt guilty about. What happened, does this event still impact you currently, and have you forgiven yourself?
- **Depression:** Describe a situation where you felt very low or depressed. What happened to make you feel that way?
- **Hurt Feelings:** Talk about a time in which someone hurt your feelings deeply. What led up to this event, how did they make you feel, and what did you do in response?

In both conditions, participants responded to a topic discussing regret, but the topic language was changed between conditions. Specifically, we removed the second question of the journal respond to since we wanted each topic to only include one question. As such, we did not include the 'regret' topic for comparison between the journal and chatbot conditions due to this change. The changed language is underlined:

- **Regret (journal version):** If you were to never see a close friend or family member again, what would you most regret not having told them? <u>Why haven't you told them yet?</u>
- **Regret (chatbot version):** If you were to never see a close friend again, what would you most regret not having told them?

In only the chatbot condition, we added a new topic for discussion:

- **Childhood (chatbot only):** How close and warm is your family? Do you feel your childhood was happier than other's?

1.2   Custom GPT-4 Internal Prompt

The GPT-4 API allowed us to define a system message which can specify the persona used by the model in its replies. Through pilot testing, we developed this custom system message, which was known by the AI chatbot, but not participants, for each conversation:

> *You are an empathic and therapeutic chatbot with your primary function being to facilitate dialogue. When users share their feelings, concerns, and challenges, try to ask them reflect and explore their emotions more deeply. Empathy is your guiding principle. Engage users as if they were confiding in a trusted therapist, and always prioritize their emotional well-being.*
>
> *The user will initiate the conversation based on a prompt. Your role is to engage in a productive dialogue for the user.*

2       Supplementary Results

2.1     Controlling for covariates in the journal-chatbot comparison

We showed in the manuscript that the happiness effect of AI conversations versus journaling differed by topic, which was ranked by the average journaling happiness. We then controlled for gender, age, education, and depression severity to ensure our findings were not influenced by covariate differences between conditions. A generalized linear mixed-effects model revealed that the relevant interaction between condition and topic rank remained significant ($\beta = 1.69 \pm 0.26, t = 6.57, p < .001$) even after controlling for these covariates (Supplementary Table 1). Interestingly, we found an interaction between topic rank and age. Specifically, older participants' happiness ratings were less affected by the topic rank. This indicates that as topics transitioned from positive (rank 1: gratitude) to negative (rank 12: hurt feelings), older adults happiness ratings decreased less than younger adults. Furthermore, significant main effects indicated that males reported lower overall happiness, and higher depression severity strongly predicted lower happiness.

**Supplementary Table 1.** Condition is a binary variable with journaling (0) and chatbot (1). Topic rank is an ordinal variable where topics are rank-ordered based on the average journaling happiness ('gratitude' is rank 1, 'hurt feelings' is rank 12). Gender is a binary variable with female (0) and male (1). Age, education, and PHQ-9 are continuous variables, z-scored across conditions. The model includes subject-specific random intercepts and random slopes for Topic rank.

**The difference in happiness between chatbot conversations and journaling was greater for more negative topics, even controlling for covariates.**

|  | **Estimates** | | | |
|---|---|---|---|---|
| *Predictors* | *Estimates* | *std. Error* | *Statistic* | *p* |
| Intercept | 72.84 | 1.63 | 44.59 | **<0.001** |
| Condition [chatbot] | -2.09 | 2.26 | -0.93 | 0.355 |
| Topic rank | -2.61 | 0.17 | -15.38 | **<0.001** |
| Gender [male] | -4.58 | 2.19 | -2.09 | **0.036** |
| Age | 1.35 | 1.06 | 1.27 | 0.204 |
| Education | 0.03 | 0.71 | 0.04 | 0.969 |
| PHQ-9 | -10.65 | 1.09 | -9.74 | **<0.001** |
| Condition [chatbot]:Topic rank | 1.69 | 0.26 | 6.57 | **<0.001** |
| Gender [male]:Topic rank | 0.40 | 0.25 | 1.61 | 0.108 |
| Age:Topic rank | -0.30 | 0.12 | -2.42 | **0.016** |
| Education:Topic rank | 0.02 | 0.12 | 0.14 | 0.886 |
| PHQ-9:Topic rank | -0.11 | 0.12 | -0.94 | 0.348 |
| N $_{sub}$ | 458 | | | |

## 2.2 Robustness checks for AI chatbot happiness boost

To ensure the robustness of the happiness increased observed with AI chatbot conversations, we conducted further analyses. Each participant in the chatbot condition responded to three randomly selected prompts. We then categorized these conversations as the 'best', 'middle', or 'worst' based on the expected happiness ratings for those topics in the journaling condition. This classification was determined by the average journaling happiness scores for each topic, independent of the participant's actual post-chatbot happiness rating. For example, if a participant discussed 'romance', 'future goals', and 'guilt', these conversations were assigned to 'best', 'middle', and 'worst' rankings, respectively. We calculated the difference between post-chatbot conversation happiness and the corresponding journaling condition happiness for each category (see Figure 2C in the manuscript). One-sample t-tests, comparing these differences to zero, revealed a significant chatbot happiness boost for 'worst' ($t(207) = 9.26, p < .001$), 'middle' ($t(207) = 6.94, p < .001$), and 'best' ($t(207) = 3.18, p = .002$) conversations. Paired t-tests further demonstrated that the chatbot happiness boost was significantly larger for 'middle' conversations compared to 'best' ($t(207) = 4.67, p < .001$), for 'worst' conversations compared to 'middle' ($t(207) = 3.46, p < .001$), and for 'worst' conversations compared to 'best' ($t(207) = 8.52, p < .001$).

To enhance the generalizability of our findings to future studies employing different topics, we categorized our prompts as either positive or negative based on the average happiness ratings from the journaling condition. Topics with an average happiness score below 50 (neutral) were classified as negative ('evaluate other', 'guilt', 'depression', and 'hurt feelings'), while those with an average above neutral were classified as positive ('gratitude', 'perfect day', 'pride', 'tv show', 'romance', 'self critical', 'future goals', and 'challenges'). Beyond the significant interaction observed between condition and topic group (as shown in Figure 2D of the manuscript), we conducted two-sample t-tests. These tests revealed a significant increase in happiness within the chatbot condition compared to the journaling condition for both positive ($t(421.05) = 3.32, p < .001$) and negative topics ($t(421.05) = 7.06, p < .001$).

## 2.3 The chatbot and journaling conditions have similar initial responses

To determine if the condition influenced participants' initial responses, we compared the first message written in the chatbot condition to the entire journaling response. Specifically, we examined word count and sentiment. Word count was calculated using tokenization from the tidytext package in R[1], after removing stop words (e.g., a, the, and). Two-sample t-tests, with corrections for multiple comparisons, revealed no significant differences in word count between the chatbot and journaling groups across all topics (all p > .18; Supplementary Figure 1A). Similarly, we conducted utterance-level sentiment analysis to calculate sentiment scores for both journal entries and the first response by participants in the chatbot condition. Two-sample t-tests, again with multiple comparisons correction, showed no significant differences in sentiment between conditions for any topic (all p > .93; Supplementary Figure 1B).

**Supplementary Figure 1. Comparisons between journal entries and the first response in the chatbot condition. A)** Word count. Comparison of word count for initial responses in the AI chatbot (orange) and journaling (red) conditions. Emotional topics are ordered by average journaling happiness. **B)** Sentiment rating. Comparison of sentiment ratings for initial responses in the AI chatbot and journaling conditions. The x-axis represents sentiment, ranging from very negative (0) to very positive (10), with a dashed line indicating the neural point (5). All statistical comparisons between chatbot and journal were non-significant (p > .05), so no significance stars are shown.

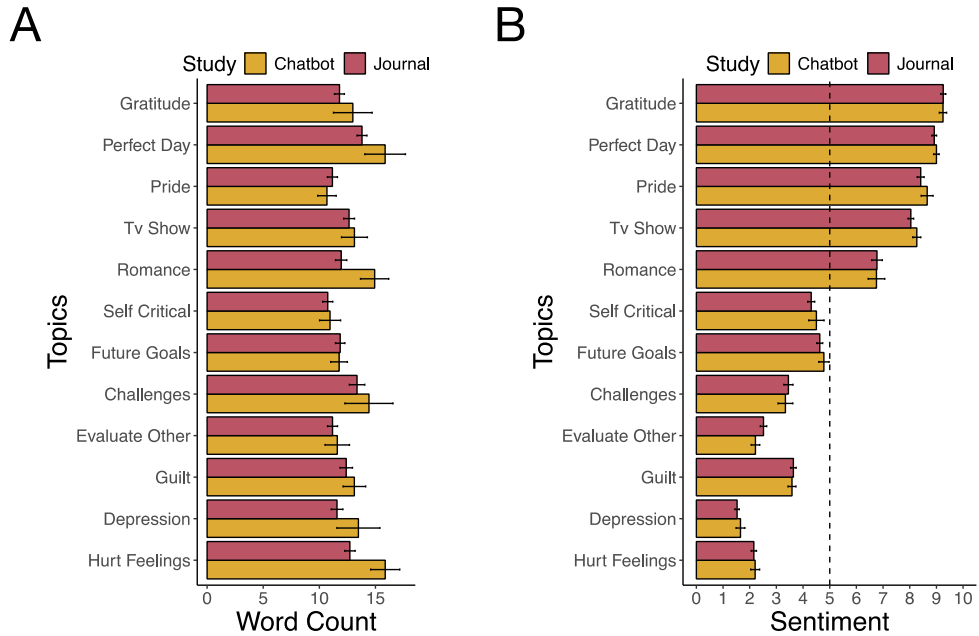

2.4     Model comparisons for the sentiment prediction error model of happiness

We used large language models to calculate sentiment scores for each utterance within a conversation. We then modeled happiness as a function of sentiment prediction errors (sPEs), defined as the difference between a participant's expected sentiment and the sentiment expressed in their utterances. Since directly asking for sentiment expectations would disrupt the conversational flow, we used the midpoint of the sentiment scale (neutral = 5) as an unbiased reference point for the initial sPE. For subsequent sPE calculations, we used the sentiment of the participant's first response to the topic as their expectation. For instance, if a participant began a 'gratitude' conversation with an utterance expressing a sentiment of 7, their first sPE would be +2 (7 – 5). Subsequent sPEs would then be calculated as the difference between each utterance's sentiment and 7 (utterance sentiment – 7). This approach ensures that subsequent responses are evaluated relative to the participant's initial sentiment expressed. To prevent participants with longer conversations from disproportionately influencing our analyses, we simplified the conversation data by assigning separate weights to the first, middle, and last utterance sPEs. For example, in a six-utterance conversation, we calculated all sPEs as described above, then extracted the first sPE, averaged the sPEs from utterances 2-5 (middle), and used the sPE from the final utterance (last). As detailed in the manuscript, when we modeled happiness as a function of the first, middle, and last sPEs using mixed-effects generalized linear regression, we found that changes in happiness were primarily predicted by the first and last sPEs (Supplementary Figure 2).

**Supplementary Figure 2. Happiness predicted by sentiment prediction errors (sPEs).** sPEs (expressed vs. expected sentiment) were categorized as first, middle, and last. The graph shows beta coefficients from a regression predicting happiness based on sPEs. ***p<.001.

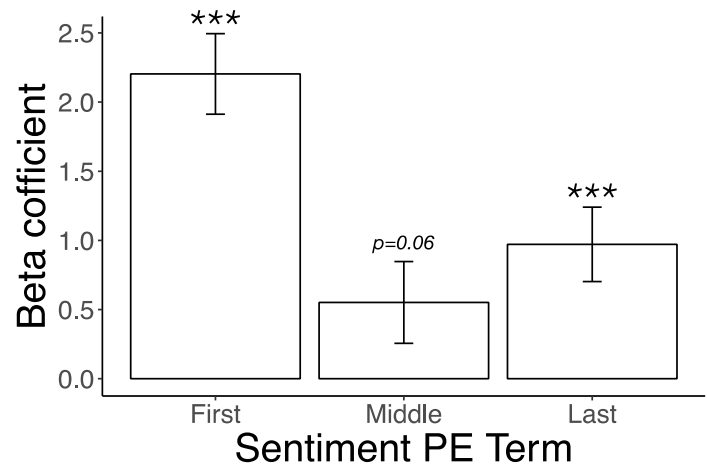

To validate this model and compare it to an alternative, we fit parameters to happiness ratings in only the chatbot condition using three-fold cross-validation. For each fold, we trained a mixed-effects generalized linear regression model on two out of three conversations per participant and then applied it to the held-out conversation for testing. Cross-validation ensures that model comparison is done using out-of-sample predictions for each model. For the alternative model, we tested a model that only applied a uniform sPE weight across all utterances, meaning it did not differentiate between first, middle, or last sPE. Model comparison using RMSE showed that our original sPE model (23.14) had lower error than the alternative model (RMSE: 24.18).

## 2.5 Sentiment increases within a conversation (normalized for length) for both chatbot and users

We investigated how AI chatbot and user sentiment changed within a conversation, limiting our analysis to the first six utterance pairs (excluding about 5% of utterance pairs). To account for varying conversation lengths, we normalized utterance pairs from 0% (conversation start) to 100% (conversation end) and interpolated the sentiment scores for intermediate utterances. Results demonstrated that sentiment increased throughout the conversation for both users ($\beta = 1.05 \pm 0.11$, t = 9.96, p < .001) and AI chatbots ($\beta = 1.39 \pm 0.11$, t = 13.09, p < .001; Supplementary Figure 3). Furthermore, a significant interaction between role (user and AI chatbot) and utterance pair was observed ($\beta = 0.33 \pm 0.13$, t = 2.53, p = .01), indicating that sentiment increased more rapidly for AI chatbots than for users across the normalized conversation timeline.

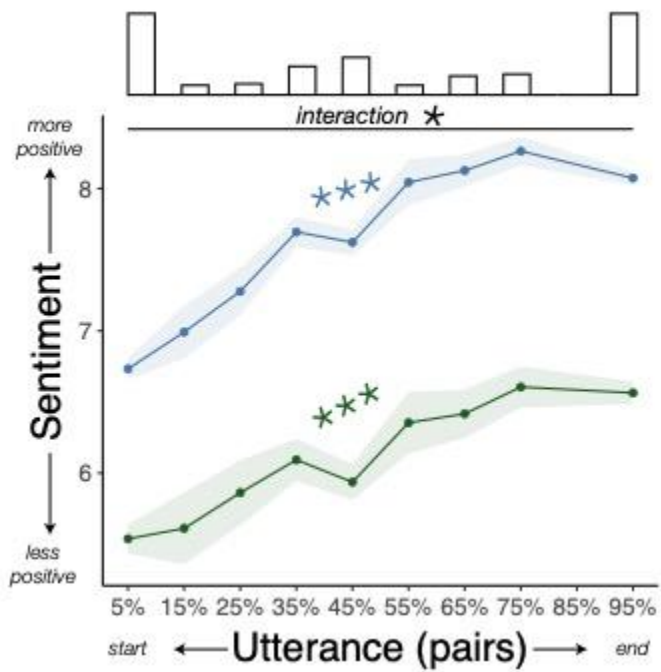

**Supplementary Figure 3. Sentiment increases within normalized conversations for both chatbot and users.** Here, 'users' refers to the participants in the AI chatbot conversations. The x-axis represents the normalized utterance pair depicting conversation progression from start to end through alternating user and chatbot responses. For visualization, normalized utterances were binned into ten equal intervals and placed at the midpoint of the bin. The y-axis displays

sentiment scores for each utterance. The histogram at the top illustrates the relative frequency of normalized utterance pairs.

**Supplementary References**

1	Silge, J. & Robinson, D. tidytext: Text Mining and Analysis Using Tidy Data Principles in R. *The Journal of Open Source Software* **1** (2016). https://doi.org:10.21105/joss.00037